\documentclass{article}

\usepackage{PRIMEarxiv}

\usepackage[utf8]{inputenc} 
\usepackage[T1]{fontenc}    
\usepackage{hyperref}       
\usepackage{url}            
\usepackage{booktabs}       
\usepackage{amsfonts}       
\usepackage{nicefrac}       
\usepackage{microtype}      
\usepackage{lipsum}
\usepackage{fancyhdr}       
\usepackage{graphicx}       
\usepackage{natbib}
\graphicspath{{media/}}     

\title{Jochre 3 and the Yiddish OCR corpus
}

\author{
  Assaf Urieli \\
  Joliciel Informatique \\
  Foix, France\\
  \texttt{assaf@joli-ciel.com} \\
   \And
  Amber Clooney, Michelle Sigiel, Grisha Leyfer \\
  Yiddish Book Center \\
  Amherst, Massachusetts\\
  \texttt{\{aclooney, msigiel, gleyfer\}@yiddishbookcenter.org} \\
}

\begin{document}
\maketitle

\begin{abstract}
We describe the construction of a publicly available Yiddish OCR Corpus, and describe and evaluate the open source OCR tool suite Jochre 3, including an Alto editor for corpus annotation, OCR software for Alto OCR layer generation, and a customizable OCR search engine. The current version of the Yiddish OCR corpus contains 658 pages, 186K tokens and 840K glyphs. The Jochre 3 OCR tool uses various fine-tuned YOLOv8 models for top-down page layout analysis, and a custom CNN network for glyph recognition. It attains a CER of 1.5\% on our test corpus, far out-performing all other existing public models for Yiddish. We analyzed the full 660M word Yiddish Book Center with Jochre 3 OCR, and the new OCR is searchable through the Yiddish Book Center OCR search engine.
\end{abstract}


\section{Introduction}
High quality OCR remains a challenge for historical documents with lower print or scan quality. This problem is exacerbated for non-standardized languages such as Yiddish, because of the variety of spelling conventions, as well as diacritics creating a very large number of potential glyphs.
Recent deep neural network approaches require a large amount of data, often produced via data augmentation: generating fake historical images from available online digital text.
However, generating such data is a challenge for Yiddish, because of the variety of spelling conventions as well as other reasons described in Section \ref{sec:yiddish}.

In 2016, the Yiddish Book Center opened its initial \href{https://ocr.yiddishbookcenter.org}{OCR search engine}\footnote{\url{https://ocr.yiddishbookcenter.org}}, enabling search of the entire Steven Spielberg Digital Yiddish Library \citep{gasztold2015yiddishBookCenter} (referred to below as the YBC Digital Library), containing over 12,000 books and over 660M words.
This website gets about 100K searches every year, from approximately 1200 unique users.
This search engine used OCR generated by Jochre 2 \citep{urieli2013jochre}, which is based on ad-hoc bottom-up page layout analysis algorithms and multinomial logistic regression for glyph recognition.
Although Jochre 2 claimed a CER of 3\% on its test corpus, it became clear over time that the error rate was much higher on actual historical texts, due to layout issues as well as non-standard orthography.

We thus decided to develop Jochre 3, a new version of Jochre using neural networks for both page layout analysis and glyph recognition.
The challenges met by Jochre 3 include learning layout from the corpus, learning to ignore diacritics, handling multiple spelling conventions and handling multiple alphabets.
In order to train and evaluate Jochre 3, we annotated a purpose-built corpus.

In Section \ref{sec:yiddish} of this article, we present some challenges of the Yiddish language with respect to OCR.
In Section 3 we present existing OCR engines and models for Yiddish.
In Section 4 we describe the Yiddish Book Center OCR corpus.
In Section 5 we describe the Jochre 3 OCR software.
In Section 6 we present our evaluation results, and compare our results with other publicly available OCR software packages for Yiddish.
Section 7 provides our conclusions.

\section{The Yiddish Language: a challenge for OCR}
\label{sec:yiddish}

In the present study, we are particularly interested in Yiddish books printed between 1880 and 1960 in the Eastern Yiddish dialect, comprising the majority of the YBC Digital Library, and will therefore concentrate on the particularities of this period.

The Yiddish language has been spoken by Ashkenazi Jews in Central and Eastern Europe since at least the 10\textsuperscript{th} century \citep{weinreich2008history,schafer2023yiddish,katz2008yiddish}.
At its apex, it was spoken by approximately 13 million speakers.
It is a synthesis of medieval German dialects with elements from Hebrew and Aramaic, as well as Slavic elements added as the Jews migrated Eastward.
Although the Germanic element is predominant, Yiddish is written in the Hebrew alphabet.

\begin{figure}[ht]
  \centering
  \includegraphics[width=10cm]{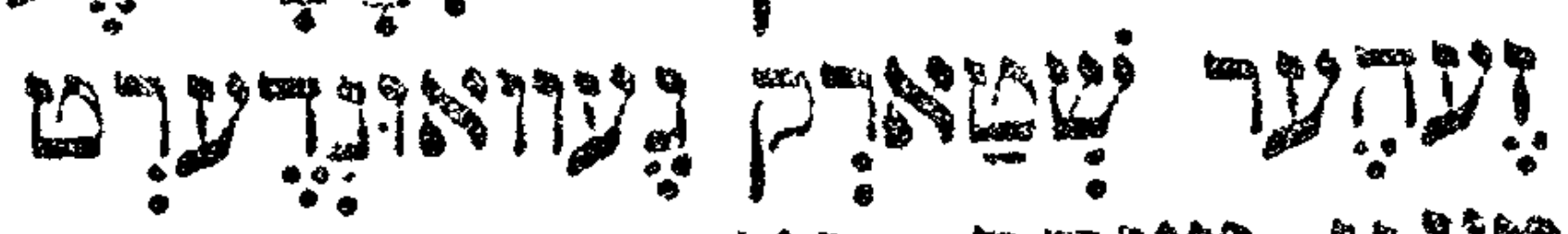}
  \caption{19\textsuperscript{th} century spelling example: ``zeyer shtark gevundert''}
  \label{fig:19thCenturySpelling}
\end{figure}

Yiddish was never the official language of a sovereign state, and as such, it does not have a single accepted orthography. Several spelling conventions exist: the YIVO spelling convention starting in 1935 \citep{yivo1935orthography}, the Soviet spelling convention starting in 1920 \citep{estraikh1999soviet}, as well as the more common ``daytshmerish'' (German influenced) orthographies used by many historical publishers.
In most orthographies, words of Hebraic and Aramaic origin are written identically to the original versions in the Bible or Talmud, with diacritics (known as ``nikud'') optionally added to indicate the vowels, whereas words of Germanic or Slavic origin are written phonetically, with certain letters from the Hebrew alphabet indicating the vowels.
19\textsuperscript{th} century publishers often combined the two systems, doubling the vowels with Hebrew diacritics, either under the vowels themselves or, more commonly, under the preceding consonant (see Figure \ref{fig:19thCenturySpelling}).

\begin{figure}[ht]
  \centering
  \includegraphics[width=10cm]{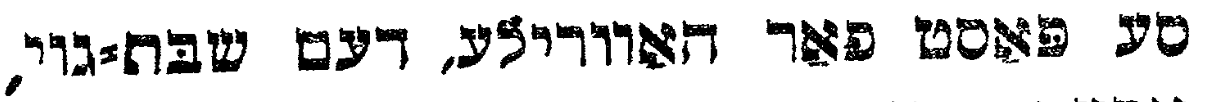}
  \includegraphics[width=10cm]{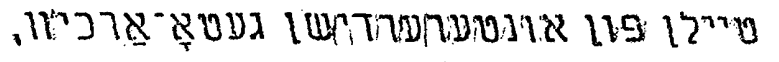}
  \includegraphics[width=10cm]{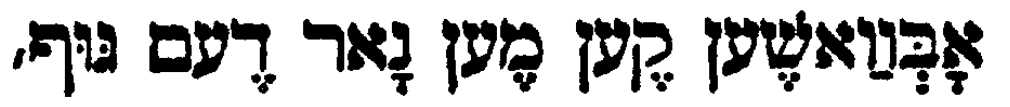}
  \includegraphics[width=10cm]{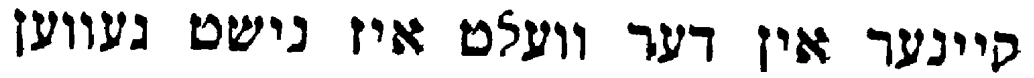}
  \includegraphics[width=10cm]{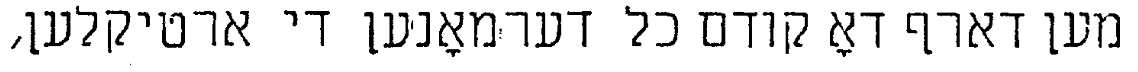}
  \caption{A sample of historic fonts in the YBC Digital Corpus}
  \label{fig:HistoricFonts}
\end{figure}

Contemporary Yiddish is mostly divided into two communities, with very little overlap between them \citep{nove2018hasidic}: the Hasidic communities, where Yiddish is spoken as an everyday language by large portions of the population, and the Yiddishist communities, mostly affiliated with academic studies of Yiddish, and highly influenced by the YIVO recommendations in terms of both spelling and vocabulary.
Hasidic Yiddish can be considered as a new dialect, very different from any of the written forms in pre-World War II Yiddish in terms of both grammar and vocabulary \citep{belk2020hasidic,belk2020loshn}.
The orthography is non-standard, partially inspired by historical daytshmerish orthography \citep{assouline2018haredi}.
This means that digital Yiddish materials available on the Internet for potential data augmentation are either in the YIVO orthography, or in a non-standardized Hasidic orthography and dialect, neither of which reflect the vast majority of historical material in the YBC Digital Library.
Equally important perhaps is the lack of accurate historical Yiddish fonts (see Figure \ref{fig:HistoricFonts}), in which artificially augmented data could be displayed.

\begin{figure}[ht]
  \centering
  \includegraphics[width=10cm]{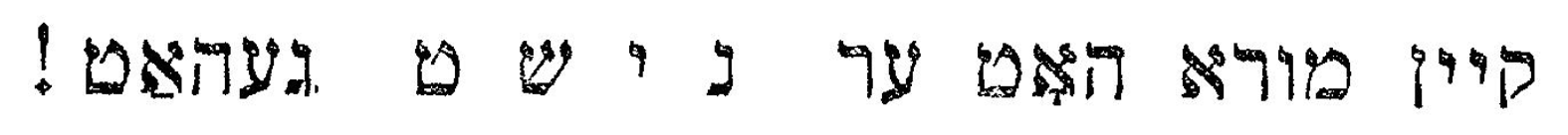}
  \caption{Spacing for emphasis in a 1920 publication: ``keyn moyre hot er \textit{nisht} gehat!''}
  \label{fig:SpacingForEmphasis}
\end{figure}

Other challenges include the common early 20\textsuperscript{th} century practice of using spacing for emphasis, instead of italics or a different font (see figure \ref{fig:SpacingForEmphasis}).
Furthermore, 19\textsuperscript{th} and 20\textsuperscript{th} century Yiddish text, especially non-fiction, can combine several alphabets, with the majority of the text written in the Hebrew alphabet, but some expressions and named entities written in the Latin alphabet (typically either in English, Polish, German or Latin) or Cyrillic alphabet (typically in Russian).
Another complication is added by the fact that, although Yiddish is written in the Hebrew alphabet from right to left, numbers are written from left to right.
This means that number strings need to be reversed post-OCR, since all modern text display systems are aware of this rule and automatically display numbers in reverse order.

These many reasons combined make OCR for 19\textsuperscript{th} and 20\textsuperscript{th} century printed Yiddish OCR, above and beyond that of the standard historical OCR challenges of print and scan quality.

\section{Existing OCR engines for Yiddish}
\label{sec:ocr-techniques}

The first OCR engine specifically trained for Yiddish was developed by Raphael Finkel.
This resulted in several important books being made available online\footnote{\url{https://www.cs.uky.edu/~raphael/yiddish.html}}. In an online interview, Finkel explains that each book requires preparatory work lasting up to several days to configure the OCR engine for the specifics of this book\footnote{\url{https://www.yiddishbookcenter.org/collections/oral-histories/interviews/woh-fi-0000599/raphael-refoyl-finkel-2014}}.

In terms of open source software, one system providing a Yiddish model out-of-the-box is Tesseract \citep{smith2007tesseract}, an OCR engine initially developed between 1984 and 1994 at HP, which introduced LSTM neural network models starting in 2018 with release 4\footnote{\url{https://tesseract-ocr.github.io/tessdoc/tess4/NeuralNetsInTesseract4.00.html}}.
These LSTM models attempt to recognize an entire line of text at a time, and require a large amount of training data.
For example, the English model required 400K text lines covering approximately 4500 different fonts.

Another open source OCR system providing a Yiddish model is Transkribus \citep{muehlberger2019transkribus}, a web platform designed for handwritten text recognition (HTR), but which can also be used for printed text.
On the Transkribus website, several published models can handle Yiddish.
Rusinek et al propose the \texttt{DiJeSt 2.0} model\footnote{\url{https://app.transkribus.org/training/text/46003}} for printed Jewish languages, including Yiddish.
\citet{reshef2024vaybertaytsh} propose the \texttt{Vaybertaytsh.YidTakNL} model, specifically constructed to recognize Vaybertaytsh, a Yiddish script in use from the 16\textsuperscript{th} to the early 19\textsuperscript{th} century.
\citet{abeliovich2022Dybbuk} proposes the \texttt{Dybbuk} model for handwritten Yiddish.
Of these, we will only evaluate the DiJeST model, as we are concerned with printed Yiddish in non-Vaybertaytsh fonts.

One very promising direction for modern OCR is found in systems such as TrOCR \citep{li2023trocr} combining a pre-trained ViT-style transformer model such as \citet{dosovitskiy2020image} for image features with a pre-trained language model such as BERT \citep{devlin2018bert} or RoBERTa \citep{liu2019roberta}) for semantic features.
In order to adapt TrOCR to Yiddish, one would need a Bert-style language model.
The only such model currently available is a multi-lingual model, XLM-RoBERTa \citep{conneau2019unsupervised}, where Yiddish is one of 100 languages.
Yiddish represents much less than 1\% of the data, and because of the limited vocabulary size, even common Yiddish words are broken up into multiple tokens.
Moreover, TrOCR requires a huge amount of data: their pretrained model for English handwriting was build on 684M lines of text.

Although \citet{kulick2023partofspeech} have recently demonstrated that an XLM-RoBERTa model fine-tuned using 660M words of ``dirty'' OCRed text in Yiddish can give promising results in POS-tagging, our own tests using this model to train TrOCR yielded very poor results on the Yiddish OCR Corpus.

We make the hypothesis here that, given the large quantity of representative data required to train LSTM or transformer-based systems for full text line recognition, one would need to generate large quantities of augmented Yiddish data.
As explained in Section \ref{sec:yiddish}, this is challenging both because contemporary Yiddish text available online is not at all representative of the YBC Digital Library, neither in spelling, nor in grammar, nor in vocabulary, and because accurate historical fonts are not currently available for many of the fonts in the corpus.

Instead, we make the assumption that high quality historical Yiddish OCR is currently best achieved by the construction of a small annotated corpus representative of the variety of the genres, spelling conventions and fonts found in the YBC Digital Library, and the application of simpler neural network models concentrating on page layout analysis on the one hand and on accurate glyph recognition on the other hand.
This led us to the construction of the Yiddish OCR Corpus, described in the next section.

\section{The Yiddish OCR Corpus}
\label{sec:ocr-corpus}

Jochre 2 was originally trained on approximately 60 pages from the YBC Digital Library.

Over 8 years of usage, users submitted many types of pages with poor OCR quality.
OCR quality was affected by the following reasons:
\begin{itemize}
    \item unusual layout, including poetry, drama, multi-column prose, dictionaries, tables of content, and indexes;
    \item a variety of historical fonts, as shown in Figure \ref{fig:HistoricFonts};
    \item unusual spelling, and in particular 19\textsuperscript{th} century works with diacritics under consonants;
    \item works containing non-Hebrew alphabet elements.
\end{itemize}

We thus decided to compile a new, much larger OCR corpus, covering all of these aspects.

The Yiddish OCR corpus\footnote{\url{https://gitlab.com/jochre/corpora/jochre-yiddish-corpus}} was annotated in the Library of Congress Alto 4 XML standard\footnote{\url{https://www.loc.gov/standards/alto/}} down to the glyph level.
The initial Alto 4 layer was generated using Jochre 2, and this layer was corrected by a 3-person annotation team using the Jochre Alto Editor\footnote{\url{https://gitlab.com/jochre/jochre-alto-editor}}.
Each page was first annotated by one team member, and then reviewed by another team member.
A separate set of 3 pages was created to measure inter-annotator agreement using Krippendorf's $\alpha$ coefficient \citep{hayes2007krippendorff} as implemented by \citet{meyer2014dkpro}.
As a distance function, we calculated CER to be the Levenshtein distance divided by the maximum string length between the two annotations.
The $\alpha$ inter-annotator agreement is 0.977 if all diacritics are taken into account, and 0.992 if the text is simplified to only YIVO diacritics (used for evaluation below).

The full annotation guideline is available together with the corpus.
In particular, we decided to annotate glyphs as closely as possible to the originals, rather than normalizing at annotation time.
In the case of dashes, for example, four types of glyphs are typically found: the standard n-dash, the standard m-dash, the Hebrew \textit{maqaf}, and a slanted equal sign.
The corpus maintains all four of these variants.
We also decided to maintain all diacritics at annotation-time, and only normalize down to a smaller set during training/evaluation.

\begin{table}[ht]
    \centering
    \begin{tabular}{|c|c|c|c|}
        \hline
         & \textbf{Training} & \textbf{Test} & \textbf{Total} \\
        \hline
        \textbf{Pages} & 554 & 104 & 658 \\
        \hline
        \textbf{Lines} & 21K & 4K & 25K \\
        \hline
        \textbf{Words} & 157K & 29K & 186K \\
        \hline
        \textbf{Glyphs} & 707K & 133K & 840K \\
        \hline
    \end{tabular}
    \caption{Yiddish OCR Corpus v1.0.0 size}
    \label{tab:corpus_size}
\end{table}

\begin{table}[ht]
    \centering
    \begin{tabular}{|c|r|r|r|}
        \hline
        \textbf{Decade} & \textbf{Pages} & \textbf{Percentage} & \textbf{CER} \\
        \hline
        \textbf{1890's} & 12 & 1.8\% & 3.95\% \\
        \hline
        \textbf{1900's} & 35 & 5.3\% & 3.02\% \\
        \hline
        \textbf{1910's} & 78 & 11.9\% & 3.24\% \\
        \hline
        \textbf{1920's} & 109 & 16.6\% & 1.49\% \\
        \hline
        \textbf{1930's} & 115 & 17.5\% & 0.96\% \\
        \hline
        \textbf{1940's} & 100 & 15.2\% & 0.98\% \\
        \hline
        \textbf{1950's} & 131 & 19.9\% & 1.13\% \\
        \hline
        \textbf{1960's} & 51 & 7.8\% & 1.17\% \\
        \hline
        \textbf{1970's} & 13 & 2.0\% & 1.15\% \\
        \hline
        \textbf{1980's} & 14 & 2.1\% & 4.40\% \\
        \hline
    \end{tabular}
    \caption{Yiddish OCR Corpus v1.0.0 distribution by decade of publication}
    \label{tab:corpus_distribution_decades}
\end{table}

\begin{table}[ht]
    \centering
    \begin{tabular}{|c|r|r|r|}
        \hline
        \textbf{City} & \textbf{Pages} & \textbf{Percentage} & \textbf{CER} \\
        \hline
        \textbf{New York} & 326 & 49.5\% & 1.73\% \\
        \hline
        \textbf{Warsaw} & 128 & 19.5\% & 1.87\% \\
        \hline
        \textbf{Buenos Aires} & 82 & 12.5\% & 0.81\% \\
        \hline
        \textbf{Vilnius} & 55 & 8.4\% & 2.37\% \\
        \hline
        \textbf{Montreal} & 34 & 5.2\% & 0.42\% \\
        \hline
        \textbf{Kyiv} & 14 & 2.1\% & 0.95\% \\
        \hline
        \textbf{London} & 12 & 1.8\% & 0.45\% \\
        \hline
        \textbf{Bucharest} & 7 & 1.1\% & 0.50\% \\
        \hline
    \end{tabular}
    \caption{Yiddish OCR Corpus v1.0.0 distribution by city of publication}
    \label{tab:corpus_distribution_city}
\end{table}

\begin{table}[ht]
    \centering
    \begin{tabular}{|c|r|r|r|}
        \hline
        \textbf{Genre} & \textbf{Pages} & \textbf{Percentage} & \textbf{CER} \\
        \hline
        \textbf{Reference} & 156 & 23.7\% & 2.77\% \\
        \hline
        \textbf{Fiction} & 148 & 22.5\% & 1.39\% \\
        \hline
        \textbf{Non-fiction} & 142 & 21.6\% & 1.13\% \\
        \hline
        \textbf{Poetry} & 118 & 17.9\% & 0.93\% \\
        \hline
        \textbf{Drama} & 52 & 7.9\% & 1.37\% \\
        \hline
        \textbf{Autobiography} & 36 & 5.5\% & 1.15\% \\
        \hline
        \textbf{Religious} & 6 & 0.9\% & 5.60\% \\
        \hline
    \end{tabular}
    \caption{Yiddish OCR Corpus v1.0.0 distribution by genre}
    \label{tab:corpus_distribution_genre}
\end{table}

The corpus size at v1.0.0 in terms of pages, words and glyphs is given in Table \ref{tab:corpus_size}.
The corpus distribution is given in Table \ref{tab:corpus_distribution_decades} by decade of publication, in Table \ref{tab:corpus_distribution_city} by city of publication, and in Table \ref{tab:corpus_distribution_genre} by genre.
The different genres enable the system to learn different page layouts, whereas the time and place of publication give a good coverage of fonts and orthographies.
The ``reference'' genre contains dictionaries, encyclopedias, collections of jokes, expressions or biographical notices and an English learning book for immigrants; ``non-fiction'' covers history, essays and travel writing; ``autobiography'' includes both diaries and memoirs; while ``religious'' is a single 19\textsuperscript{th} book containing the biography and teachings of a rabbi.
We list CER evaluation results as well in the interest of space, but will only discuss these results in the section on evaluation (Section \ref{sec:evaluation} below).
Annotation time, which depended to a great extent on the quality of the uncorrected OCR, was approximately 30 minutes per page.
The corpus is available under a Creative Commons Attribution-NonCommercial-ShareAlike license.

\section{The Jochre 3 OCR software}
\label{sec:jochre}

Jochre stands for Java Optical CHaracter REcognition, although the latest version is actually written in Scala rather than Java.
The current version, Jochre 3\footnote{\url{https://gitlab.com/jochre/jochre3-ocr}}, performs both Page Layout Analysis and Text Recognition, as well as providing a separate OCR Search Engine. It is available under an Affero GPL license.

\subsection{Page Layout Analysis}

In terms of page layout analysis, we first exported the Yiddish OCR Corpus to YOLO \citep{redmon2016YOLO} object detection format.
Before exporting, all pages were converted to grey-scale, deskewed (we assumed skewing to be limited to rotation), normalized for brightness/contrast, and reduced to a resolution of 1280$\times$1280.
We exported the following object types separately (always excluding the test corpus):
\begin{itemize}
    \item Top-level blocks: composed blocks (typically columns of text), standalone text blocks not contained in any composed block, and illustrations. We decided not to attempt to detect separators for now (e.g. lines between main text and footnotes), nor do we attempt to distinguish between content text blocks and reference material, such as page numbers or running titles. This is the only export which is not limited to the print area, and thus includes margins. It is also the only export which distinguishes two classes: text and illustrations.
    \item Text blocks: all text blocks. According to our annotation guide, a text block represents a group of text lines followed by a hard newline. This implies a single paragraph in prose, and a single line of verse in poetry (which can sometimes span across two physical text lines for long lines of verse). This export is only useful for identifying paragraphs and other hard newlines in the generated OCR.
    \item Text lines: a physical line of text, represented as a rectangle centered on the base-line, and with a thickness equal to 1\% of the page height.
    \item Words: groups of glyphs separated from other groups of glyphs by white space. This includes words with attached punctuation. Space-emphasized words are considered as a single word.
    \item Glyphs: individual glyphs. Yiddish is ambiguous with respect to single glyphs in the case of ligatures: ``tsvey vovn'' (two ``vov'' letters), ``tsvey yudn'' (two ``yud'' letters), and ``vov yud''. After experimentation, best results were attained by considering all ligatures as two separate glyphs except for ``pasekh tsvey yudn'' (two ``yud'' letters with a single ``pasekh'' diacritic beneath both of them). Before exporting glyphs and reducing resolution, the print area was divided into 4 separate tiles, each overlapping the neighboring tiles by 1/8\textsuperscript{th} of the page height/width. Tiling increases the relative size of glyph rectangles, which vastly improves glyph detection accuracy.
\end{itemize}

We then fine-tuned a YOLOv8 pre-trained object detection model for each of these object types, with 10\% of the images set aside for model validation.
We used the pre-trained ``nano'' model for all object types except for top-level blocks, which used the pre-trained ``small'' model.
Strangely, the top-level block model, although trained at a resolution of 1280$\times$1280, gave significantly better results when evaluating on images reduced to 640$\times$640.

When performing detection, Jochre first converts the page to grey-scale, deskews it, normalizes its brightness and contrast, and reduces its resolution.
It then applies each of the object type models to the image, resulting for each object type in a set of rectangular objects each associated with a confidence score.
The top-level blocks are used to determine the print area, before detecting other object types.
In case of two objects of the same type with significant overlap, Jochre either merges the two objects (top-level blocks above a certain confidence) or eliminates the lower-confidence object (all other object types).
If the overlap is slight, Jochre places the border at the half way point to remove the overlap.
It then combines the objects into a single page layout analysis by starting with the top-level blocks, and placing each lower-level object inside a single containing higher-level object.

\subsection{Text recognition}
In terms of text recognition, as mentioned in Section \ref{sec:ocr-techniques}, we decided to concentrate on a simple glyph recognition network rather than a full word or text line recognition network.
In other words, once the image has been segmented down to glyph level, the software attempts to recognize each individual glyph using a CNN model.

The model architecture, inspired by \citet{deote2021MNIST}, is:
\begin{itemize}
    \item 28$\times$28 input image (centered on the glyph, with the longer dimension normalized to 28, and no attempt to remove neighboring glyphs from the image)
    \item Convolution layer: 32 features, 3$\times$3 kernel + ReLU activation + Batch normalization
    \item Convolution layer: 32 features, 3$\times$3 kernel + ReLU activation + Batch normalization
    \item Convolution layer: 32 features, 5$\times$5 kernel, stride=2 + ReLU activation + Batch normalization. The stride of 2 replaces max pooling by sub-sampling.
    \item 40\% dropout.
    \item Convolution layer: 64 features, 3$\times$3 kernel + ReLU activation + Batch normalization
    \item Convolution layer: 64 features, 3$\times$3 kernel + ReLU activation + Batch normalization
    \item Convolution layer: 64 features, 5$\times$5 kernel, stride=2 + ReLU activation + Batch normalization. The stride of 2 replaces max pooling by sub-sampling.
    \item 40\% dropout.
    \item Flatten to a 1-dimensional vector.
    \item Dense layer of size 128 + ReLU activation + Batch normalization.
    \item 40\% dropout.
    \item Dense layer of size equal to the number of classes with softmax applied. 
\end{itemize}

A glyph recognition model is trained for each alphabet.
There is no knowledge of character n-grams: the only contextual knowledge is graphical, since parts of neighboring glyphs are included to the right and left.
The Hebrew alphabet model first normalizes the expected glyphs so that only diacritics used in YIVO Yiddish are kept, and any other diacritics are ignored.
Similarly, n-dashes are normalized to a single glyph, as is the case for several other punctuation classes.
Any Latin alphabet glyphs are assigned the character ``L'', and any Cyrillic alphabet glyphs are assigned the character ``C''.
This gives a total of 88 possible glyphs.

The Latin and Cyrillic alphabet models are trained respectively on Latin and Cyrillic glyphs only. Note that we limited our training to the Latin and Cyrillic glyphs found in the Yiddish corpus - a better model could be constructed if additional material were used for these models.

When performing text recognition, if the Hebrew alphabet model generates ``L'' or ``C'' for any letter in the word, the Latin or Cyrillic model is then applied to guess the actual glyph.

Jochre also uses a wide coverage lexicon of possible Yiddish words in YIVO spelling.
When detecting, it is possible to generate a beam of possible predictions for each word, and then re-rank predictions based on their appearance in the lexicon.
Before checking for predicted word existence, the word is first converted to YIVO spelling using a set of ad-hoc rules.
If the word does not exist in the lexicon, its total score (the geometric mean of glyph confidence scores) is multiplied by an unknown-word factor $< 1.0$.
Similarly, impossible words (e.g. with final letters in median positions) are multiplied by an impossible-word factor of 0.01.
The predicted word with the highest re-ranked score is then selected.
A beam width of 5 and unknown-word factor of 0.5 gave the best results.

\subsection{OCR Search Engine}

\begin{figure}[ht]
  \centering
  \fbox{\includegraphics[width=16cm]{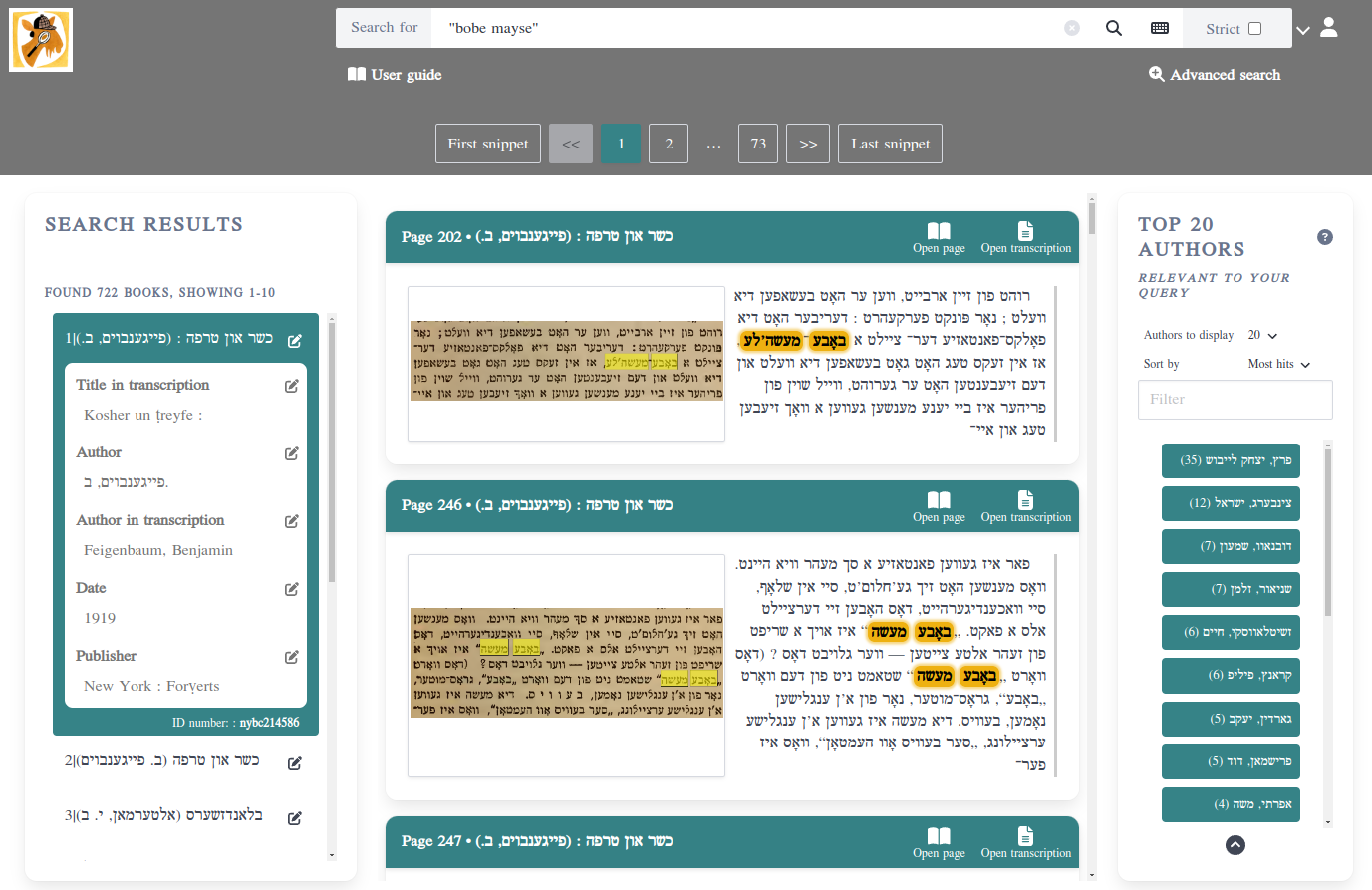}}
  \caption{OCR searching using the Jochre 3 Search Engine}
  \label{fig:Jochre3SearchEngine}
\end{figure}

The final component of Jochre 3 is an OCR search engine, including a REST API backend\footnote{\url{https://gitlab.com/jochre/jochre3-search}} and a customizable Vue.js frontend\footnote{\url{https://gitlab.com/jochre/jochre3-search-client}} (see Figure \ref{fig:Jochre3SearchEngine}).
This search engine is built on top of Apache Lucene\footnote{\url{https://lucene.apache.org/}}, and stores coordinates for all words in a separate database.
When a search is performed, it enables query extension to all words sharing the same lemma as the search term.
Using the stored word coordinates, it can also highlight terms matching the search results in the original image.
Finally, it includes functionality to allow users to crowd-source OCR corrections, as well as corrections to the book's metadata.

Note that the Jochre 3 OCR search engine is completely independent from the Jochre 3 OCR generation software.
All it requires is a PDF (or a set of page images) and corresponding Alto files produced by any OCR tool.

\section{Evaluation}
\label{sec:evaluation}

We performed an evaluation of Jochre 3 v1.0.0 on the test corpus, and compared it with results for Jochre 2 v2.6.5, the Transkribus DiJeST 2.0 model as run on 17/07/2024, the Tesseract v4.1.1 Yiddish model, and the Tesseract v4.1.1 combined Yiddish and English model.
In all cases, we first normalized both expected and predicted text to the YIVO alphabet, in order to compare like with like.
We also removed all newlines beyond those marking a physical end-of-line, to concentrate evaluation on actual text recognition rather than paragraph recognition.
We measure the following metrics, giving the mean value per page in all cases:
\begin{itemize}
    \item CER: the character error rate, calculated as the Levenshtein distance divided by the gold text length. This measure is highly sensitive to page reading order, and therefore to page layout analysis.
    \item BoW: the cosine similarity of the expected and predicted text when converted to a bag of words. This measure is not sensitive to page reading order.
    \item Time: the analysis time for a page.
\end{itemize}

Results are shown in Table \ref{tab:evaluation}.

\begin{table}[ht]
    \centering
    \begin{tabular}{|l|r|r|r|}
        \hline
        \textbf{Model} & \textbf{CER} & \textbf{BoW} & \textbf{Time} \\
        \hline
        Jochre 2 & 17.0\% & 0.854 & 3.2s \\
        \hline
        Transkribus DiJeST 2.0 & 19.9\% & 0.870 & 20.7s \\
        \hline
        Tesseract Yiddish & 9.0\% & 0.832 & \textbf{2.1s} \\
        \hline
        Tesseract Yiddish+English & 10.1\% & 0.839 & 3.7s \\
        \hline
        Jochre 3 & \textbf{1.5\%} & \textbf{0.978} & 6.3s \\
        \hline
    \end{tabular}
    \caption{Evaluation results}
    \label{tab:evaluation}
\end{table}

Jochre 3 gives by far the best OCR quality, with CER 6 times better than the nearest candidate, and the best Bag of Words cosine similarity (0.978). In terms of speed, it is about 3 times slower than Tesseract Yiddish.
The high CER for Jochre 2 can be partially explained by the fact that the corpus was chosen to represent pages that Jochre 2 had difficulty analyzing.
Time estimates for the Transkribus DiJeST model are not directly comparable to other time estimates, as the calculation was performed on the Transkribus server, rather than locally.

In terms of CER by decade (see Table \ref{tab:corpus_distribution_decades}), there is clearly a higher error rate for works prior to 1920 than for works after 1920.
This is certainly due to a more decorative layout and more wear and tear.
The outlier in the 1980's concerns a reference work with many very small footnotes.

In terms of CER by city (see Table \ref{tab:corpus_distribution_city}), there are no obvious conclusions. On the other hand, CER by genre (see Table \ref{tab:corpus_distribution_genre}) gives some obvious differences. The ``reference'' genre with a CER of 2.70\% contains dictionaries and other material with a very complex layout. Similarly, the ``religious'' genre with a CER of 5.60\% is a single 19\textsuperscript{th} century book, typical of religious books from this period, with crowded text, full diacritics, and considerable wear and tear. All other genres give error rates around 1\%.

Another less formal evaluation can be performed by searching for typical Yiddish phrases in the entire YBC corpus, as analyzed by Jochre 3 as opposed to Jochre 2.
We would expect to find more matches in the Jochre 3 version, where each match represents one book containing at least one instance of the phrase in question.
The results are shown in Table \ref{tab:identical_searches}, with on average 2.70 as many matches in Jochre 3.

\begin{table}[ht]
    \centering
    \begin{tabular}{|l|l|r|r|r|}
        \hline
        \textbf{Phrase} & \textbf{Meaning} & \textbf{Jochre 2} & \textbf{Jochre 3} & \textbf{Ratio} \\
        \hline
        ``beshum oyfn nit'' & ``absolutely not'' & 2800 & 5598 & 2.00 \\
        \hline
        ``bobe mayse'' & ``fairy tale'' & 140 & 730 & 5.21 \\
        \hline
        ``keyn eyn hore'' & ``no evil eye'' & 1044 & 2300 & 2.20 \\
        \hline
        ``kholile nit'' & ``God forbid'' & 1877 & 4302 & 2.29 \\
        \hline
        ``mishteyns gezogt'' & ``alas'' & 856 & 1548 & 1.81 \\
        \hline
    \end{tabular}
    \caption{Identical searches in Jochre 2 and 3}
    \label{tab:identical_searches}
\end{table}

\section{Conclusion}
\label{sec:conclusion}

In the present study, we built an annotated OCR corpus for Yiddish, and developed and evaluated Jochre 3, an OCR tool providing high quality OCR for Yiddish.
Our evaluation shows that Jochre 3 outperforms other OCR tools available for Yiddish by far, attaining a CER of 1.5\% on our test corpus.
This supports our hypothesis that, given the current state of Yiddish digital resources, and the differences between the historical YBC Digital Library and contemporary Yiddish material on the Internet, excellent results are attained using simple neural networks concentrating on glyph recognition.
This has the additional advantage of requiring less training material, fewer computing resources, and being much quicker to train.

Similar models could easily be built for other lesser resourced languages, especially ones with non-standard orthographies.
Constructing such models would require an OCR corpus annotated in Alto down to glyph level using the Jochre Alto editor, and possibly some configuration to simplify the alphabet and deal with multi-alphabet cases or other language peculiarities (e.g. left-to-right numbers in a right-to-left alphabet).
Once such models are constructed, the Jochre 3 search engine could easily be used to expose the OCRed text to the public.

Although our current system is limited to page layout analysis and glyph recognition, it would be very interesting to attempt to construct more complex models which take into account semantics in order to correct the OCR.
Future directions could include constructing a BERT-style Yiddish language model out of the 660M word YBC Digital Library, now that the OCR quality has been vastly improved.
This Yiddish language model could either be used to perform OCR post-correction, or directly incorporated into OCR analysis in a TrOCR style model.
However, in order to generate sufficient OCR training data for a TrOCR style model, we would have to overcome two preliminary issues: graphic work would be required to reconstruct historical Yiddish fonts, and software development would be required to convert contemporary online Yiddish material to historical spelling conventions.

The high-quality Yiddish OCR Corpus constructed in the present study will make it possible to perform a reliable evaluation and thus measure future improvements in Yiddish OCR tools and models.

As it stands, the full YBC Digital Library as analyzed by Jochre 3, with much higher OCR quality than before, is now available at \url{https://ocr.yiddishbookcenter.org}.
This corpus and associated search engine, covering over 12,000 historical books, provide a highly valuable resource for researchers and students of Yiddish worldwide.
The Jochre 3 tool is also being used in the Universal Yiddish Library (UYL) project currently underway (planned for beta release in March 2025), which aims at providing OCR search capabilities to the combined digital Yiddish libraries of the Yiddish Book Center, the National Library of Israel, the New York Public Library, and the YIVO Institute for Jewish Research.

\section*{Acknowledgments}
This study was supported by the Yiddish Book Center and Joliciel Informatique.
A special thanks to Mirjam Gutschow for her excellent review and comments.

\bibliographystyle{agsm}  
\bibliography{references}  

\end{document}